\documentclass{article}

\usepackage[preprint]{corl_2026}

\usepackage{amsmath}
\usepackage{amssymb}
\usepackage{mathrsfs}

\usepackage{amsthm}

\usepackage{xcolor}
\usepackage{hyperref}
\usepackage{subcaption}
\usepackage{booktabs}

\usepackage{macros}

\title{Learning from Imperfect Demonstrations via Temporal Behavior Tree-Guided Trajectory Repair}

\author{
  Aniruddh G.~Puranic\thanks{Equal Contribution}\\
  Institute of Systems Research (ISR)\\
  University of Maryland, College Park\\
  United States\\
  \texttt{puranic@umd.edu} \\
  \And
  Sebastian Schirmer$^*$\\
  DLR German Aerospace Center \\
  Department of Unmanned Aircraft, Braunschweig\\
  Germany\\
  \texttt{sebastian.schirmer@dlr.de} \\
  \AND
  John S.~Baras \\
  Institute of Systems Research (ISR)\\
  University of Maryland, College Park\\
  United States\\
  \texttt{baras@umd.edu} \\
  \And
  Calin Belta \\
  Institute of Systems Research (ISR)\\
  University of Maryland, College Park\\
  United States\\
  \texttt{calin@umd.edu} \\
}

\begin{document}
\maketitle


\begin{abstract}
    Learning robot control policies from demonstrations is a powerful paradigm, yet real-world data is often suboptimal, noisy, or otherwise imperfect, posing significant challenges for imitation and reinforcement learning. In this work, we present a formal framework that leverages Temporal Behavior Trees (TBT), an extension of Signal Temporal Logic (STL) with Behavior Tree semantics, to repair suboptimal trajectories prior to their use in downstream policy learning. Given demonstrations that violate a TBT specification, a model-based repair algorithm corrects trajectory segments to satisfy the formal constraints, yielding a dataset that is both logically consistent and interpretable. The repaired trajectories are then used to extract potential functions that shape the reward signal for reinforcement learning, guiding the agent toward task-consistent regions of the state space without requiring knowledge of the agent's kinematic model. We demonstrate the effectiveness of this framework on discrete grid-world navigation and continuous single and multi-agent reach-avoid tasks, highlighting its potential for data-efficient robot learning in settings where high-quality demonstrations cannot be assumed.
\end{abstract}


\section{Introduction}

Learning control policies from demonstrations is a powerful paradigm for training robot controllers without hand-crafted reward functions. However, real-world demonstrations often suffer from multiple challenges: they may be suboptimal, contain noise, violate desired specifications, or be incomplete~\cite{ravichandar_recent_2020}. These imperfections pose significant challenges to imitation and reinforcement learning, potentially leading to policies that fail safety-critical requirements or exhibit poor generalization in novel scenarios. Rather than discarding such imperfect data or requiring extensive manual curation, we propose a formal and principled framework that \emph{repairs} demonstrations prior to their use in downstream learning. By combining Temporal Behavior Trees (TBT)~\cite{schirmer2024_tbt}, an expressive formal specification language grounded in Signal Temporal Logic (STL), with a model-based repair algorithm, we transform specification-violating demonstrations into logically consistent, specification-satisfying trajectories that serve as a supervisory signal for policy learning.

Our approach operates in three stages: (1) \emph{trajectory repair}, where TBT specifications and a model-based repair procedure correct trajectory segments that violate formal task constraints, yielding a cleaned dataset that is formally verifiable; (2) \emph{reward discovery}, where potential functions are extracted from the repaired trajectories to construct dense reward signals for reinforcement learning (RL), reducing the need for manual reward engineering; and (3) \emph{policy learning}, where the potential-based reward and the TBT online monitor~\cite{matheu2025_omtbt} are used to train a robust policy on the environment.

A central advantage of the proposed framework is that it preserves interpretability and structural guarantees. By operating within the formal logic of TBTs, the learned policies inherit well-defined semantics and can be subject to formal verification. Furthermore, since the potential functions are extracted purely from demonstrated state trajectories in pose space, the reward shaping signal is independent of the agent's specific kinematic model, making the framework directly applicable across robotic platforms obeying different kinematic models (e.g., Ackermann, Dubins, or unicycle dynamics), without modification to the reward shaping pipeline. \emph{The key insight is that formal trajectory repair produces specification-consistent state sequences that serve as a kinematic-model-agnostic supervisory signal for policy learning, bridging the gap between formal methods and data-driven robot learning without requiring access to actions, rewards, or agent dynamics.} To the best of our knowledge, this is the first work to apply TBT-based trajectory repair~\cite{schirmer2025_repairtbt} to downstream policy learning, extending its utility beyond specification verification to data-driven control synthesis.

\begin{figure*}
    \centering
    \includegraphics[width=\textwidth]{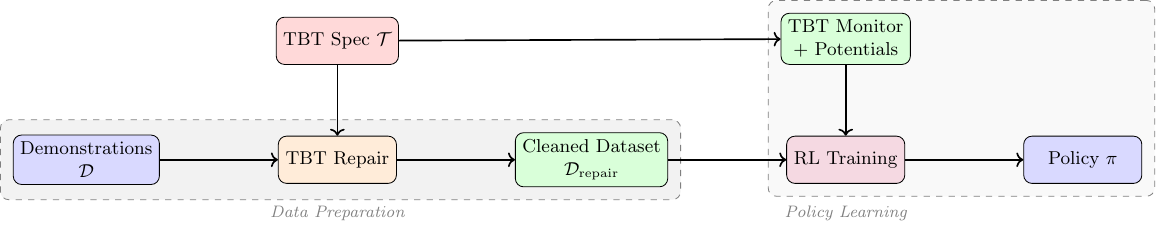}
    \caption{Overview of the proposed framework. The TBT specification $\tree$ governs both the repair of suboptimal demonstrations and the construction of the TBT monitor and reward potentials used during RL training.}
    \label{fig:framework}
\end{figure*}

The contributions of this paper are as follows:
\begin{itemize}
    \item We propose a formal framework for policy learning that integrates TBT-based trajectory repair with RL, enabling policy acquisition from imperfect demonstrations without assuming access to optimal data or accurate agent kinematic models.

    \item We demonstrate how repaired trajectories can be used to extract potential functions for potential-based reward shaping~\cite{ng_policy_invariance}, providing a dense and specification-consistent guidance signal for RL that is agnostic to the agent's kinematic model.

    \item We instantiate the framework in a discrete grid-world navigation setting, where explicit potential functions derived from repaired trajectories accelerate RL convergence across map sizes of varying complexity, and validate it on continuous, high-dimensional single and multi-agent reach-avoid tasks, where the proposed approach matches or exceeds the performance of expert-designed reward functions while incurring significantly lower obstacle cost rates.
\end{itemize}

\section{Related Work}
\label{sec:related}

The introduction of TBTs~\cite{schirmer2024_tbt} enabled trace segmentation, i.e., a robustness-based analysis of trajectories. BT2Automata~\cite{matheu2025_bt2automata} extended this by translating BTs to Timed Automata, supporting formal verification, consistency checking, and correct-by-construction synthesis. OMTBT~\cite{matheu2025_omtbt} further developed quantitative semantics for partial BT executions, enabling real-time monitoring and logic-based reward shaping for RL. Relevant to this work, Schirmer et al.~\cite{schirmer2025_repairtbt} developed optimization-based trace repair methods that adjust observed behaviors to satisfy TBT specifications using robust semantics and trajectory segmentation. Collectively, these works form a coherent framework spanning specification, verification, monitoring, and repair of robotic behaviors.

Several works have leveraged the quantitative semantics of STL to define or shape reward functions for RL and imitation learning~\cite{aksaray_qlearning_2016, li_reinforcement_2017, anandb_rl, jiang2020temporallogicbased}. Formal methods have also been used to support learning from demonstration: \cite{belta_lfd} combines a logic-augmented automaton with an MDP to initialize RL from expert demonstrations, while \cite{wenchao_safety_al} introduces a counterexample-guided approach for safety-aware apprenticeship learning. Other works integrate logic specifications directly into policy learning via loss functions~\cite{InnesRSS20} or model-predictive control~\cite{cho_mpc_stl}. Co-safe LTL has been used to jointly learn automata and reward functions from demonstrations~\cite{wen_lfd_highlevel}, though such approaches often assume optimal demonstrations and suffer from exponential state-space growth. Most closely related, \cite{PDN_ALSTL} and \cite{PDN21} evaluate suboptimal demonstrations via STL quantitative semantics to infer reward functions for RL with STL monitoring. In contrast, our framework leverages the more expressive TBT logic to repair suboptimal trajectories and extract specification-consistent potential functions for downstream policy learning.

\section{Preliminaries}
\label{sec:prelims}

We model the interaction between the agent and environment as a Markov Decision Process.

\begin{definition}[Markov Decision Process]
    A Markov Decision Process (MDP) is a tuple $(\StateSpace, \ActionSpace, T, I, R)$, where: $\StateSpace \subset \Reals^k$ is the state space of the system and $\ActionSpace \subset \Reals^l$ is its action space. The actions can be discrete or real-valued (continuous), and can be finite or infinite. $T: \StateSpace \times \ActionSpace \times \StateSpace \rightarrow [0, 1]$ is the transition function (a probability distribution). It maps a transition $(s, a, s')$ to a probability, i.e., $T(s, a, s') = Pr(s_{t+1}=s' \mid s_t=s, a_t=a)$. $I$ is the distribution of the initial state $s_0$. $R$ is a reward function that typically maps a state $s \in \StateSpace$, or a transition $(s, a, s') \in \StateSpace \times \ActionSpace \times \StateSpace$ to $\Reals$.
\label{def:mdp}
\end{definition}

An MDP includes an optional discount factor denoted by $\gamma \in [0, 1]$. In RL, the goal of the learning algorithm is to find a policy $\pi: \StateSpace \times \ActionSpace \rightarrow [0,1]$ that maximizes the total (discounted) reward from performing actions on an MDP. The objective is to maximize $\sum_{t = 0}^{\infty}{\gamma^t r_t}$, where $r_t = R(s_t)$ or $R(s_t, a_t, s_{t+1})$, $s_0 \sim I(s_0)$, $a_t \sim \pi(a_t | s_t)$ and $s_{t+1} \sim T(s_t, a_t, s_{t+1})$. We assume full observation of the state space for agents operating in known environments.

\begin{definition}[Trajectory or Episode Rollout]
    A trajectory in an MDP is a sequence of state-action pairs of finite length $L \in \mathbb{N}$ obtained by following some policy $\pi$ from an initial state $s_0$. A trajectory is denoted as: $\tau =\langle(s_i, \pi(s_i))\rangle_{i=0}^{L-1}$, where $s_i \in \StateSpace$ and $a_i \in \ActionSpace$. 
    \label{def:rollout}
\end{definition}

To express that an agent achieves its learning goal, we check if its trajectory $\tau$ satisfies the goal specification $\tree$, denoted as $\tau \models \tree$. The goal specification is given as Temporal Behavior Tree $\tree$, which allows expressing sequential and parallel task properties in a tree fashion.

\begin{definition}
    [Syntax of Temporal Behavior Trees \cite{schirmer2024_tbt}]
    We construct a TBT $\tree$ using the following syntax:
     \[\def\arraystretch{1.0} \begin{array}{rll}
    \tree & ::=  \Sequence( [\tree, \tree]), & \leftarrow \text{Sequence node} \\
    & |\ \Parallel_M( [\tree, \dots, \tree]), M \in \mathbb{N} & \leftarrow \text{Parallel node}\\
    & |\ \Fallback([\tree, \dots, \tree]), & \leftarrow \text{Fallback node} \\
    & |\ \Leaf(\varphi), & \leftarrow \text{Leaf node} \\ 
    \end{array}\]
    where $\varphi$ is a local property expressed in a temporal language.
    Note that we restrict our focus to a selected subset of TBT operators for brevity and clarity. 
\end{definition}

Informally, a sequence of tasks in which a subtask $\tree_1$ must be satisfied before another subtask $\tree_2$ is represented by $\Sequence( [\tree_1, \tree_2])$. For convenience, we rewrite $\Sequence([\tree_1, (\Sequence([\tree_2, \cdots, \Sequence( [\tree_{n-1}, \tree_n]))))$ for $n \geq 2$ as $\Sequence([\tree_1, \ldots, \tree_n])$. A parallel composition of $n$ subtasks, where at least $M$ subtasks must be satisfied, is expressed by $\Parallel_M( [\tree_1, \dots, \tree_n])$. Further, given multiple subtasks where the satisfaction of any single subtask suffices, this is expressed by $\Fallback(\tree_1, \dots, \tree_n)$. Last, the objective of an individual task is specified at a leaf node by a temporal formula $\varphi$, i.e., $\Leaf(\varphi)$. 

For expressing local properties $\varphi$, we use Signal Temporal Logic (STL). STL is a real-time logic, generally interpreted over a dense-time domain for signals whose values are from a continuous metric space (such as $\Reals^n$). The basic primitive in STL is a {\em signal predicate} $\mu$ that is a formula of the form $f(\vx(t)) > 0$, where $\vx(t)$ is the tuple $(s,a)$ of the trajectory $\vx$ at time $t$, and $f$ maps the signal domain $\domain = (\StateSpace \times \ActionSpace)$ to $\Reals$. STL formulas are then defined recursively using Boolean combinations of sub-formulas, or by applying an interval-restricted temporal operator to a sub-formula. The syntax of STL is formally defined as follows: $\varphi ::=  \mu \mid \neg \varphi \mid \varphi \wedge \varphi \mid \alw_{I} \varphi  \mid \ev_{I} \varphi \mid \varphi \until_{I} \varphi$. Here, $I = [a,b]$ denotes an arbitrary time-interval, where $a,b\in\Reals^{\ge 0}$. In this work, the semantics of STL are defined over a discrete-time signal $\sig$ defined over some time-domain $\timedomain$. The Boolean satisfaction of a signal predicate is simply \textit{True} ($\top$) if the predicate is satisfied and \textit{False} ($\bot$) if it is not, the semantics for the propositional logic operators $\neg, \land$ (and thus $\lor, \rightarrow$) follow the obvious semantics. The following behaviors are represented by the temporal operators:
\begin{itemize}
    \item At any time $t$, $\always_I(\varphi)$ says that $\varphi$ must hold for all samples in $t+I$.
    \item At any time $t$, $\eventually_I(\varphi)$ says that $\varphi$ must hold \textit{at
    least once} for samples in $t+I$.
    \item At any time $t$, $\varphi \until_I \psi$ says that $\psi$ must hold at some time $t'$ in $t+I$, and in $[t,t')$, $\varphi$ must hold at all times.
\end{itemize}

The quantitative (robustness) semantics of STL, introduced in \cite{fainekos_robustness_2009,donze_robust_2010}, provide a measure of how well trajectories satisfy a specification. Unlike Boolean semantics, which yield only a binary outcome, robustness semantics assign a numerical value. A positive value indicates that the specification is satisfied, whereas a negative value indicates violation. Further, the magnitude of this value reflects the degree of satisfaction and violation. The definitions of the robustness semantics of TBT and STL are provided below.

\begin{definition}[Robustness Semantics of TBTs]\label{def-tbt:robustness-semantics-tbt}
    The robustness of a temporal behavior tree $\tree$ on a finite execution trace $\sigma$, denoted $\rho(\tree, \sigma)$, is defined as follows:  
    \begin{align*}
    \rho(\Sequence( [\tree_1, \tree_2]), \sigma) & = \max\limits_{u\in[0,|\sigma|-1]} \min (\rho(\tree_1, \sigma[:u]), \\
    & \phantom{==} \rho(\tree_2, \sigma[u+1:]))\\
    \rho(\Parallel_M([\tree_1, \dots, \tree_k]), \sigma) & = max_M( 
    \rho(\tree_1, \sigma),\cdots, \\
    & \phantom{==} \rho(\tree_k, \sigma)) \\ 
    \rho(\Fallback( [\tree_1, \dots, \tree_k])),\ \sigma) & = \max\limits_{j\in[1,k]} \max\limits_{i\in[0,|\sigma|-1]}~ \rho(\tree_j, \sigma[i:]) \\
    \rho(\Leaf(\varphi),\ \sigma) & = \rho(\varphi, \sigma)\\
    \end{align*}
\end{definition}

\begin{definition}[STL Robust Semantics]\label{def-tbt:robustness-stl}
    Robustness of an STL formula $\varphi$ over a trace $\sigma$, denoted $\rho(\varphi, \sigma)$, is defined as follows:
    \begin{align*}
    \rho(p_i,\ \sigma) & = f_i(\sigma(1)) ~\mbox{, if}~ |\sigma| > 0 ~\mbox{else}~ -\infty\\
    \rho(\neg \varphi,\ \sigma) & = - \rho(\varphi, \sigma)\\
    \rho(\varphi_1 \land \varphi_2, \ \sigma) & = \min(\rho(\varphi_1, \sigma), \rho( \varphi_2, \sigma))\\
    \rho(\varphi_1 \lor \varphi_2,\ \sigma) & = \max(\rho(\varphi_1, \sigma), \rho( \varphi_2, \sigma))\\
    \rho(\F_{[l,u]} ( \varphi ),\ \sigma) & = \max\limits_{i \in [l,u]} ~~\rho(\varphi, \sigma[i:])\\
    \rho(\G_{[l,u]} (\varphi),\ \sigma) & = \min\limits_{i \in [l,u]} ~~\rho(\varphi, \sigma[i:])\\
    \rho(\varphi_1 \ \Un_{[l,u]}\ \varphi_2,\ \sigma) & = \max\limits_{i \in [l,u]} \min(\rho(\varphi_2,\sigma[i:]), \\
    & \phantom{=} \min\limits_{j \in [0,i-1]}  \rho(\varphi_1, \sigma[j:]))
    \end{align*}
\end{definition}

Multi-agent tasks benefit from the parallel operator in TBTs, as it avoids the combinatorial explosion of logical formulas. For example, consider a task in which any two out of three goals must be reached. With TBTs this can be expressed as: $\Parallel_2(\F \mathit{reachGoal_A},\F \mathit{reachGoal_B}, \F \mathit{reachGoal_C})$.

\section{Problem Formulation}

For an MDP as in \autoref{def:mdp}, we are given: (i) a finite dataset of demonstrations $\dataset = \{\demo_1, \demo_2, \cdots, \demo_m\}$, where each $\demo_i$ is a trajectory defined by \autoref{def:rollout}, and (ii) a specification in TBT $\tree$ that describe the task(s). A subset of $\dataset$ may not satisfy $\tree$. The objective is to infer a control policy $\pi$ for an agent such that its behavior satisfies $\tree$. Formally, our goal is to solve the following problem:
\begin{equation}
    \max_{\pi} \Pr_{\tau \sim \pi} \left[
    \tau \models \tree \right]
\end{equation}

\section{Solution}
\label{sec:theory}

To solve the above problem, we propose the framework illustrated in \autoref{fig:framework}. It comprises two modules: {\em TBT Repair} and {\em RL-based policy learning}. Given a set of demonstrations $\dataset$ and a TBT specification $\tree$ describing the task, the TBT Repair mechanism checks each trajectory in $\dataset$ against $\tree$, repairing those that violate the specification while leaving satisfying trajectories unchanged, yielding the repaired dataset $\drepair$. The cleaned dataset $\drepair$ is subsequently used by the RL module to extract a potential function that shapes the reward signal, guiding the agent toward task-consistent and safe regions of the state space. An RL agent is then trained using the shaped reward in conjunction with a TBT monitor that provides dense feedback based on $\tree$, ultimately yielding a policy $\pi$ whose behavior satisfies the specification. The following subsections describe each module in detail.

\subsection{TBT Trace Repair}
\label{sec:tbt-repair}

Repairing violating demonstrations in an optimal manner, such that they satisfy a TBT specification, can be encoded as a mixed-integer linear program (MILP)~\cite{schirmer2025_repairtbt}. Let $\Model$ denote a linear approximation of the transition dynamics $T$ of the MDP, which encodes the feasible state transitions of the agent (e.g., a Dubins vehicle model for continuous navigation or a discrete adjacency model for grid-worlds). Given $\Model$, a TBT specification $\tree$, and a trajectory $\tau$ with $\tau \not\models \tree$, we construct a repaired trajectory $\tau_R$ that satisfies $\tree$ with $|\tau_R| = |\tau|$, denoted by $\Repair(\tree, \Model, \tau)$, as follows:
\[
\begin{array}{rll}
\arg\min_{\tau_{R}} & J(\tau, \tau_R) & \leftarrow \text{Cost function}\\  
s.t. & \tau_R \models \Encode_{\tau_R}(\Model) & \leftarrow \text{Trace follows model}\\
& \tau_R \models \Encode_{\tau_R}(\tree) & \leftarrow \text{Trace satisfies TBT}
\end{array}
\]
The cost function $J(\tau, \tau_R) = \sum_{t=0}^{|\tau|-1} d(s_t, s_t^R)$ measures the total deviation of the repaired trajectory from the original, where $d: \mathcal{S} \times \mathcal{S} \rightarrow \mathbb{R}_{\geq 0}$ is a distance metric over the state space (e.g., Manhattan distance for discrete grid-worlds or Euclidean distance for continuous domains). This ensures that the repaired trajectory remains as close as possible to the original demonstration, preserving its core behavioral intent rather than introducing an alternative solution. However, computing optimal repairs is often infeasible in practice, as the underlying MILP formulation is NP-hard. Further, such optimality is not strictly required for demonstration purposes. If possible, we therefore adopt a landmark-based repair strategy \cite{schirmer2025_repairtbt} where so-called landmarks simplify the MILP formulation to a linear program (LP). A landmarks resolves disjunctions in the TBT specification in a greedy manner. Specifically, it leverages the robustness semantics. For instance, if there is a disjunction of two predicates, then the predicate with the larger robustness value is chosen as landmark. In principle, this repair resolves all binary choices the optimizer must make. As example, consider again  $\Parallel_2(\F \mathit{reachGoal_A},\F \mathit{reachGoal_B}, \F \mathit{reachGoal_C})$ and an executed trajectory $\tau$ that violates this specification. The specification contains several disjunctions: the parallel-operator must select which goal locations to reach, while each eventually-operator introduces additional disjunctions over the time step at which its corresponding goal is reached. For our landmark-based repair, we first compute the temporally closest candidate positions for satisfying two goals. These are then enforced by introducing corresponding constraints into the LP. If the optimizer fails to find a feasible solution, we iteratively test other landmarks until a predefined time bound is exceeded.

\subsection{Control Synthesis via RL}
\label{sec:rl-control}

Given the repaired/cleaned dataset $\drepair$ obtained from the TBT repair procedure, we aim to learn a control policy $\pi$ that satisfies the TBT specification $\tree$. We extract a potential function from $\drepair$ that is used to shape the reward signal for RL. This formulation has a key advantage: since the potential function is derived purely from the repaired state sequences, it is independent of the agent's specific kinematic model, making it directly applicable across different robotic platforms without modification to the reward shaping pipeline (e.g., mobile robots obeying Ackermann, Dubins, or unicycle dynamics, as demonstrated in this work).

Formally, let $\mathcal{R} = \bigcup_{i} \{\mathbf{p}_t^{(i)}\}$ denote the set of all poses pooled from the trajectories in $\drepair$, where $\mathbf{p}_t^{(i)}$ would represent the 2D or 3D pose of the agent $i$ at time $t$. We define two complementary potential functions over the agent's current pose $\mathbf{p}_t$. The first is a \textit{proximity potential} that rewards the agent for being spatially close to poses visited in these trajectories:
\begin{equation}
    \Phi_{\text{prox}}(\mathbf{p}_t) = \exp\left(-\min_{\hat{\mathbf{p}} \in \mathcal{R}} 
    \|\mathbf{p}_t - \hat{\mathbf{p}}\|_2\right).
    \label{eq:phi_prox}
\end{equation}
The second is a \textit{safety potential} that penalizes the agent for deviating beyond a threshold $\epsilon$ from the repaired trajectory poses, enforcing that the agent remains within the safe region implicitly defined by $\drepair$:
\begin{equation}
    \Phi_{\text{safe}}(\mathbf{p}_t) = -\Ind\left[\min_{\hat{\mathbf{p}} \in \mathcal{R}} 
    \|\mathbf{p}_t - \hat{\mathbf{p}}\|_2 > \epsilon\right].
    \label{eq:phi_safe}
\end{equation}
These potential functions complement each other: $\Phi_{\text{prox}}$ provides fine-grained guidance within the safe region, while $\Phi_{\text{safe}}$ acts as a coarser safety boundary that discourages large deviations. The two potentials are combined as a weighted sum to form the shaping reward:
\begin{equation}
    F(\mathbf{p}_t) = \alpha \cdot \Phi_{\text{prox}}(\mathbf{p}_t) + 
    \beta \cdot \Phi_{\text{safe}}(\mathbf{p}_t),
    \label{eq:shaping}
\end{equation}
where $\alpha, \beta \geq 0$ are scalar weights. Following the potential-based reward shaping framework of~\cite{ng_policy_invariance}, the shaped reward presented to the RL agent at each timestep is:
\begin{equation}
    r'(s_t, s_{t+1}) = r_{\text{env}}(s_t, s_{t+1}) + \gamma \cdot F(\mathbf{p}_{t+1}) - 
    F(\mathbf{p}_t),
    \label{eq:shaped_reward}
\end{equation}
where $r_{\text{env}}$ is the environment's reward based on the TBT monitor's robustness value~\cite{matheu2025_omtbt} and $\gamma$ is the discount factor used by the RL algorithm. 
By construction, this shaping term does not alter the set of optimal policies~\cite{ng_policy_invariance}, while providing a dense guidance signal that steers the agent toward task-consistent regions of the state space identified by the TBT repair procedure. Crucially, since the potential functions are derived from TBT-repaired trajectories that satisfy $\tree$ by construction, the shaping signal implicitly encodes the formal specification into the reward structure, thus promoting specification satisfaction through guidance while preserving the policy optimality guarantees of standard RL theory. Nearest neighbor queries for computing $\min_{\hat{\mathbf{p}} \in \mathcal{R}} \|\mathbf{p}_t - \hat{\mathbf{p}}\|_2$ are accelerated using a KD-tree~\cite{bentley1975multidimensional} fitted on $\mathcal{R}$ prior to training, reducing per-timestep query complexity from $\mathcal{O}(|\mathcal{R}|)$ to $\mathcal{O}(\log |\mathcal{R}|)$. Standard off-the-shelf RL is then performed with the shaped reward $r'$, with no further modifications to the learning algorithm.

\section{Experiments}
\label{sec:experiments}

We evaluate the proposed framework on two settings of increasing complexity: a discrete grid-world navigation task, and continuous single and multi-agent reach-avoid tasks. All experiments were conducted on a machine equipped with an Intel Core Ultra 9 285K (24-core) processor, 64GB RAM, and an NVIDIA RTX 5090 GPU. The TBT repair optimization was solved using the Gurobi optimizer~\cite{gurobi}, while RL training was implemented using Stable-Baselines3~\cite{stable_baselines3}.

\subsection{Grid-World}

We consider a $10 \times 10$ grid-world environment~\cite{gym_simplegrid} consisting of a set of states $\mathcal{S} = \{\text{start}, \text{goal}, \text{obstacles}\}$, with obstacles assigned randomly. The task is described by the TBT specification $\tree \defeq \Parallel_2(\ev\alw \mathit{reachGoal}, \alw \mathit{avoidCollision})$, which requires the agent to reach and remain at the goal while avoiding collisions with obstacles. At each timestep, the observation is a binary encoding of the agent's current grid position, and the action space consists of the four orthogonal moves $\{\uparrow, \rightarrow, \downarrow, \leftarrow\}$. As the environment is deterministic, $T(s, a, s') = \Pr(s_{t+1} = s' \mid s_t = s, a_t = a) = 1$, and the Manhattan distance metric is used for nearest neighbor queries.

\begin{figure}[htbp]
    \centering
    \subfloat[$10 \times 10$ grid map\label{fig:grid-map}]{\includegraphics[width=0.3\textwidth]{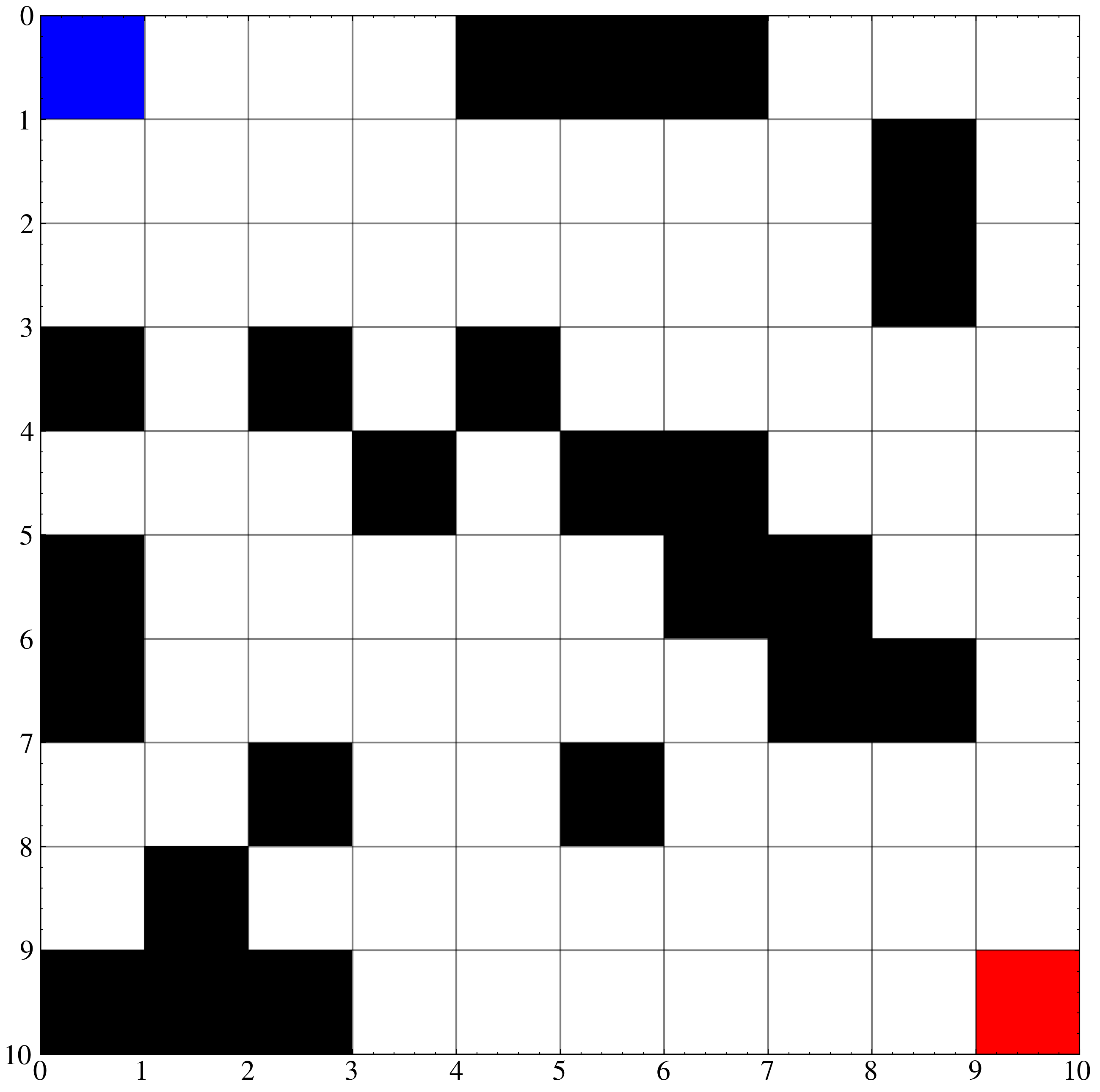}} \hfil
    \subfloat[Demonstrations\label{fig:grid-demos}]{\includegraphics[trim=0 0 0 7mm,clip,width=0.3\textwidth]{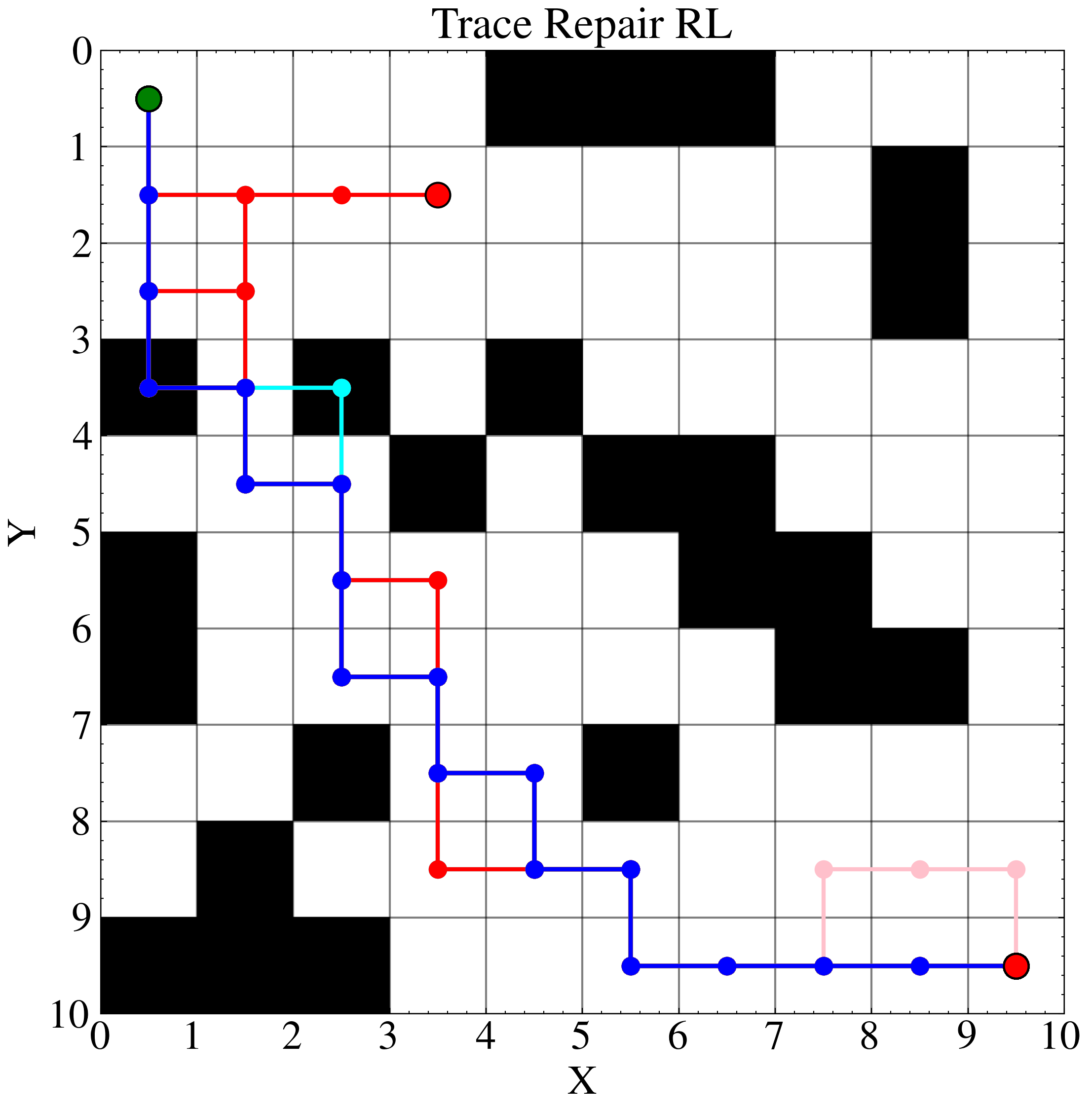}}

    \subfloat[Potential Functions\label{fig:grid-potentials}]{\includegraphics[width=0.49\textwidth]{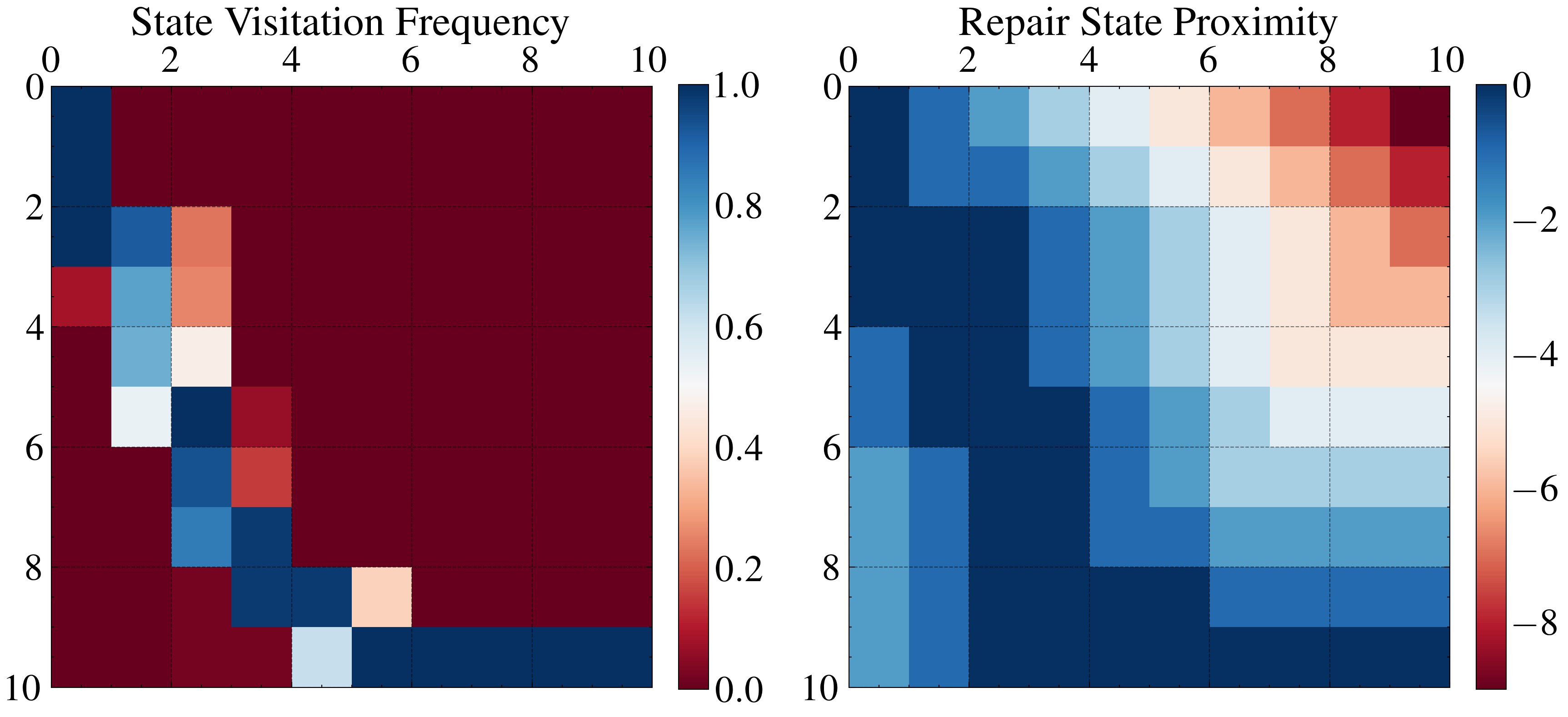}} \hfil    
    \subfloat[Performance Comparison\label{fig:grid-comparison}]{\includegraphics[width=0.49\textwidth]{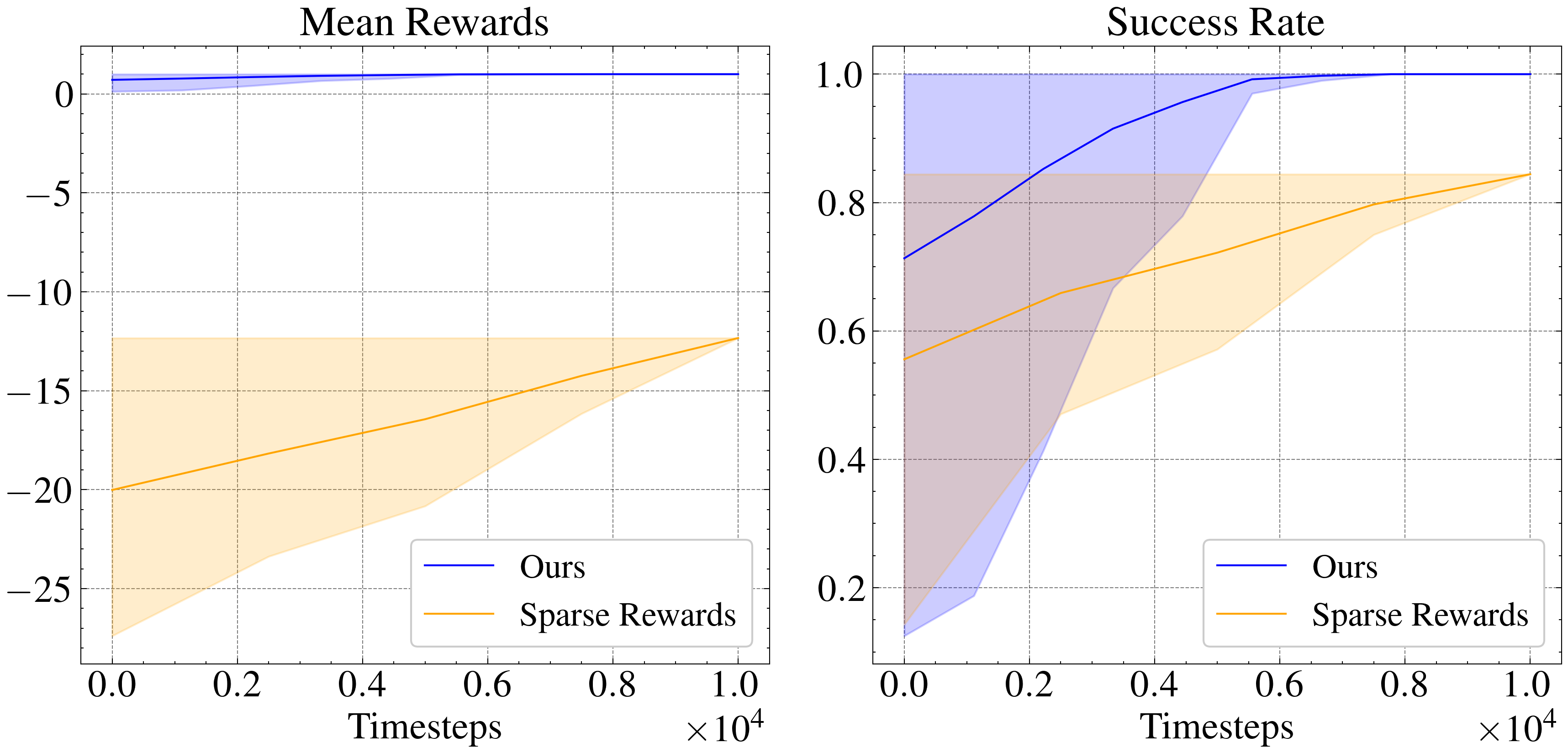}}
    
    \caption{Grid-World environment: \textbf{(a)}: Blue indicates start location at the top-left, red is the goal at the bottom-right, and the black cells are obstacles; \textbf{(b)} Example demonstrations that violate the TBT specification by visiting an unsafe cell. \textbf{(c)} Heatmaps of the extracted potential functions from the cleaned trajectories; \textbf{(d)} Comparison of the mean episodic rewards and success rates.}
    \label{fig:grid-world}
\end{figure}

We collect $50$ demonstrations by training an expert RL agent and recording trajectories at periodic checkpoints, where earlier checkpoints correspond to suboptimal or TBT-violating trajectories. These are repaired using the TBT repair tool to yield $\drepair$, with each repair completing in under $1$ second. Note that movement in the discrete grid-world does not permit the MILP to be reduced to an LP; nevertheless, a repair for violating trajectories was consistently found within this time budget. \autoref{fig:grid-world} shows the potential functions $\Phi_{\text{prox}}$ and $\Phi_{\text{safe}}$ computed from $\drepair$, alongside training curves comparing the proposed approach against a sparse reward PPO~\cite{ppo_schulman} baseline with identical hyperparameters for the $10 \times 10$ grid map. The potential heatmaps confirm that repaired trajectories encode meaningful spatial structure, wherein obstacle cells are assigned low potential while task-consistent corridors between start and goal exhibit high potential. The proposed approach converges faster and achieves higher asymptotic performance across all map sizes, with the gap most pronounced in larger maps where sparse rewards provide an increasingly weak learning signal.

\subsection{Mobile Navigation}

\begin{figure*}[htbp]
\centering

\begin{subfigure}[b]{0.4\textwidth}
    \centering
    \includegraphics[width=0.8\textwidth,height=1.4in]{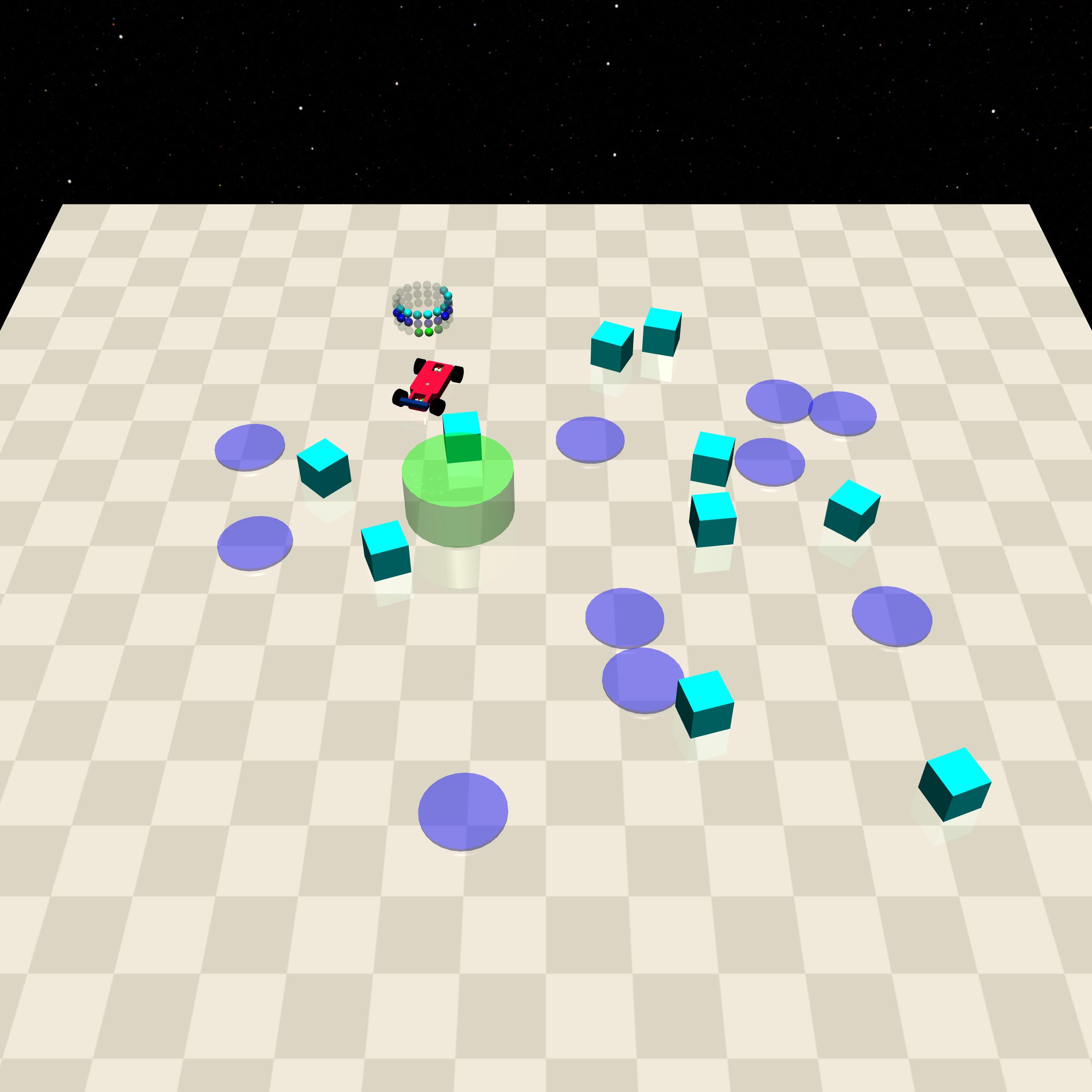}
    \caption{Single-agent task}
\end{subfigure}
\hfil
\begin{subfigure}[b]{0.4\textwidth}
    \centering
    \includegraphics[width=0.8\textwidth,height=1.4in]{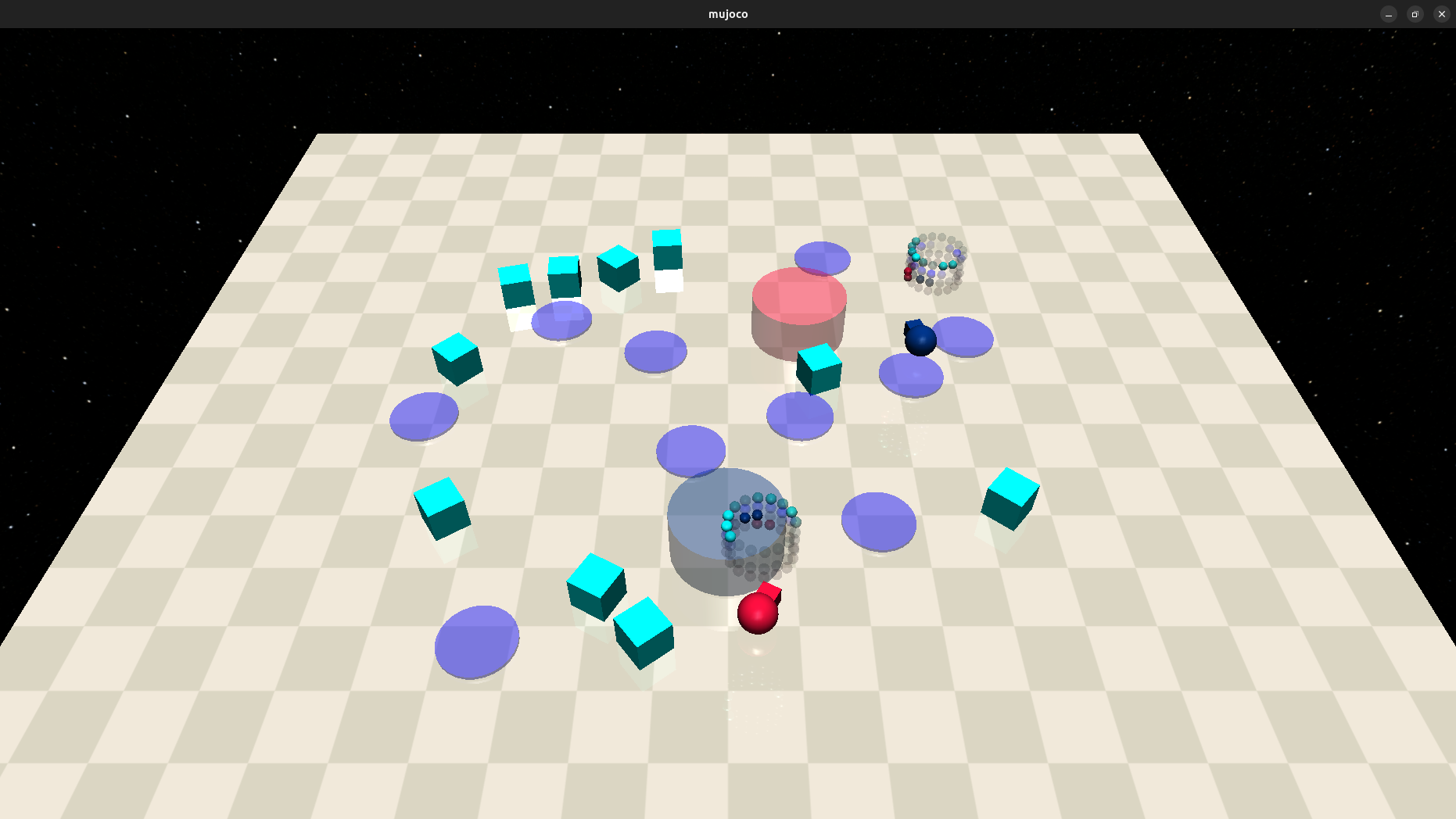}
    \caption{Multi-agent task}
\end{subfigure}

\begin{subfigure}[b]{0.32\textwidth}
    \centering
    \includegraphics[width=0.99\textwidth,height=1.2in]{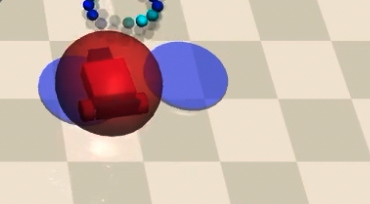}
    \caption{Restricted region traversal}
\end{subfigure} \hfil
\begin{subfigure}[b]{0.32\textwidth}
    \centering
    \includegraphics[width=0.99\textwidth,height=1.2in]{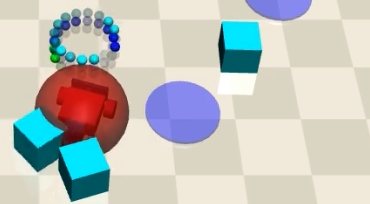}
    \caption{Obstacle collision}
\end{subfigure} \hfil
\begin{subfigure}[b]{0.32\textwidth}
    \centering
    \includegraphics[width=0.99\textwidth,height=1.2in]{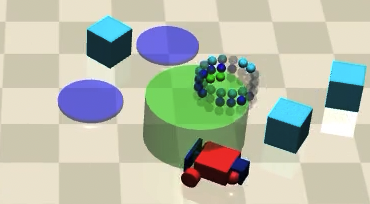}
    \caption{Goal reached (success)}
\end{subfigure}

\caption{Reach-avoid navigation task: \textbf{(a)} A close-up of our experiment setup showing a mobile robot (indicated by a car-like object) tasked with navigating to a uniformly randomized 2D goal location (green cylinder); \textbf{(b)} Multi-agent extension of the single-agent environment. The robots are indicated by red and blue car-like objects and the goals are indicated by red and blue cylinders. \textbf{(c)--(e)} Each robot must avoid two types of obstacles: a traversable soft constraint (purple discs) and hard constraints that inhibit motion upon collision (cyan cubes). These hazards incur costs that the agent must minimize while learning to safely navigate to the goal.}
\label{fig:safety-gym-env}
\end{figure*}

We evaluate the proposed approach on two reach-avoid tasks (\autoref{fig:safety-gym-env}) in the Safety-Gymnasium environment~\cite{ji2023safetygym}, where agent(s) must navigate to goal location(s) (indicated by green cylinders) while avoiding hard (cyan cubes) and soft obstacle regions (purple discs). Traversing a purple disc or colliding with a cyan cube incurs a cost or penalty. The distance to the goal and hazards are provided by lidar measurements, and the environment contains 20 hazard markers scattered around the map. At the start of each episode, the initial locations of the agent(s), obstacles, and goal(s) are randomized. The episode is terminated if the agent reaches the goal, hits an obstacle or traverses outside the map boundary.

In both tasks, $50$ demonstrations are collected from a baseline RL agent trained with no cost penalty for obstacle collisions, and repaired using the TBT repair tool to yield $\drepair$. As movement is in a continuous space, the MILP is reduced to an LP using a linearized Dubins vehicle model\footnote{The Dubins model models car-like robots with forward-only motion and bounded curvature, the Ackermann steering geometry describes their physical steering geometry through steering angle and wheelbase, and the unicycle model abstracts mobile robots as systems with independent linear and angular velocity control.\label{fn:kinematic-models}}, with each repair completing in under $3$ seconds. Crucially, the repaired demonstrations cover only a subset of the state space, and the inferred potential function has no knowledge of the model used to generate the demonstrations. The 2D poses pooled from $\drepair$ are indexed using a KD-tree as described in \autoref{sec:rl-control}, fitted once prior to training and queried at each timestep to efficiently retrieve the nearest repaired pose to the agent's current position, ensuring that the reward shaping computation introduces negligible overhead during RL training. We compare the proposed approach against two baselines: a PPO baseline trained with an expert-designed dense reward function, and Safe-PPO, a variant of PPO that augments the standard policy network with a dedicated cost-minimization network to explicitly penalize constraint violations. In contrast, the proposed approach requires no architectural modifications and leverages off-the-shelf PPO directly. In all cases, trained models were evaluated over $100$ test episodes to record their statistics.

\subsubsection{Single-Agent Reach-Avoid}

\begin{figure}[htbp]
    \centering
    \subfloat[Single-agent\label{fig:single-comparison}]{\includegraphics[width=0.49\textwidth]{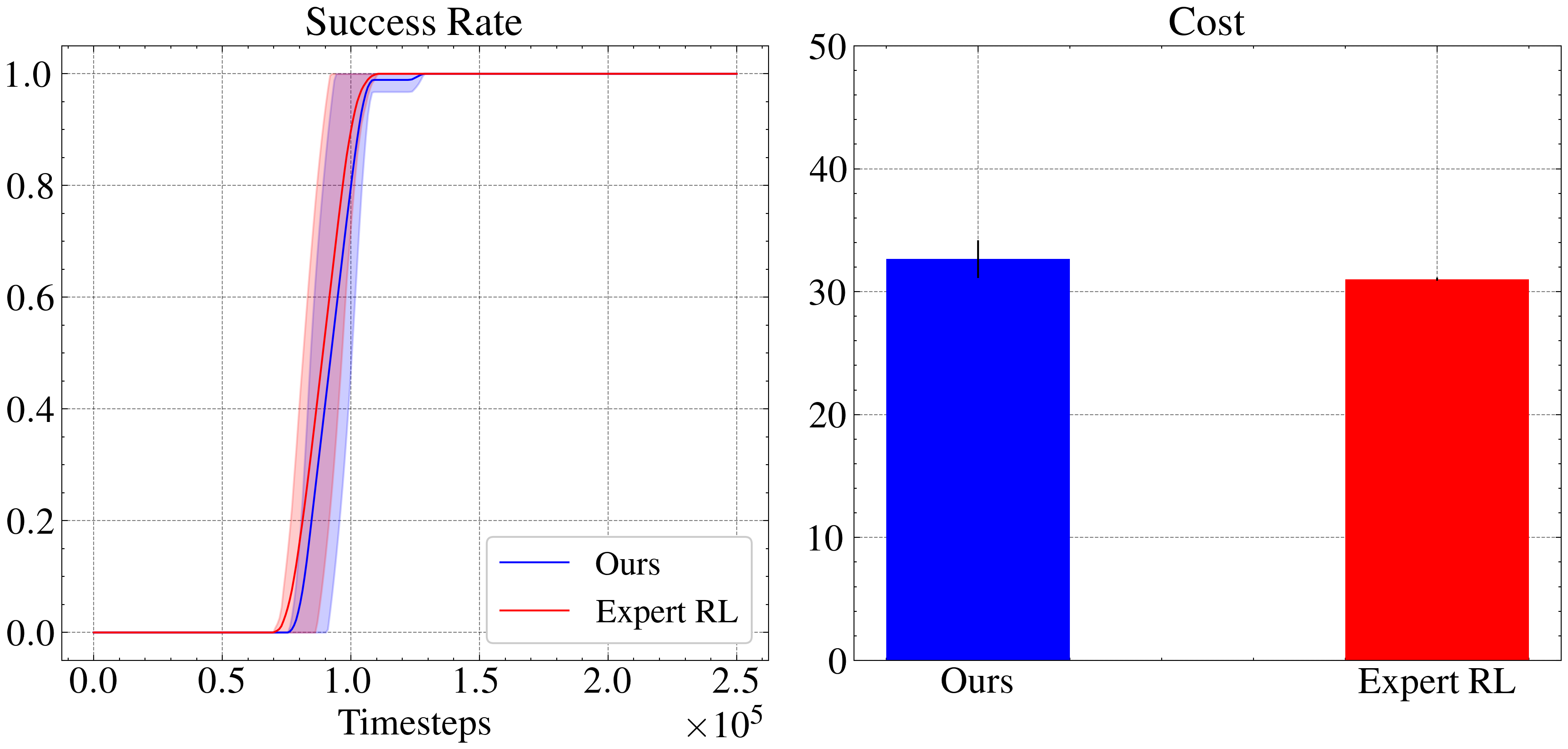}}    
    \subfloat[Multi-agent\label{fig:multi-comparison}]{\includegraphics[width=0.49\textwidth]{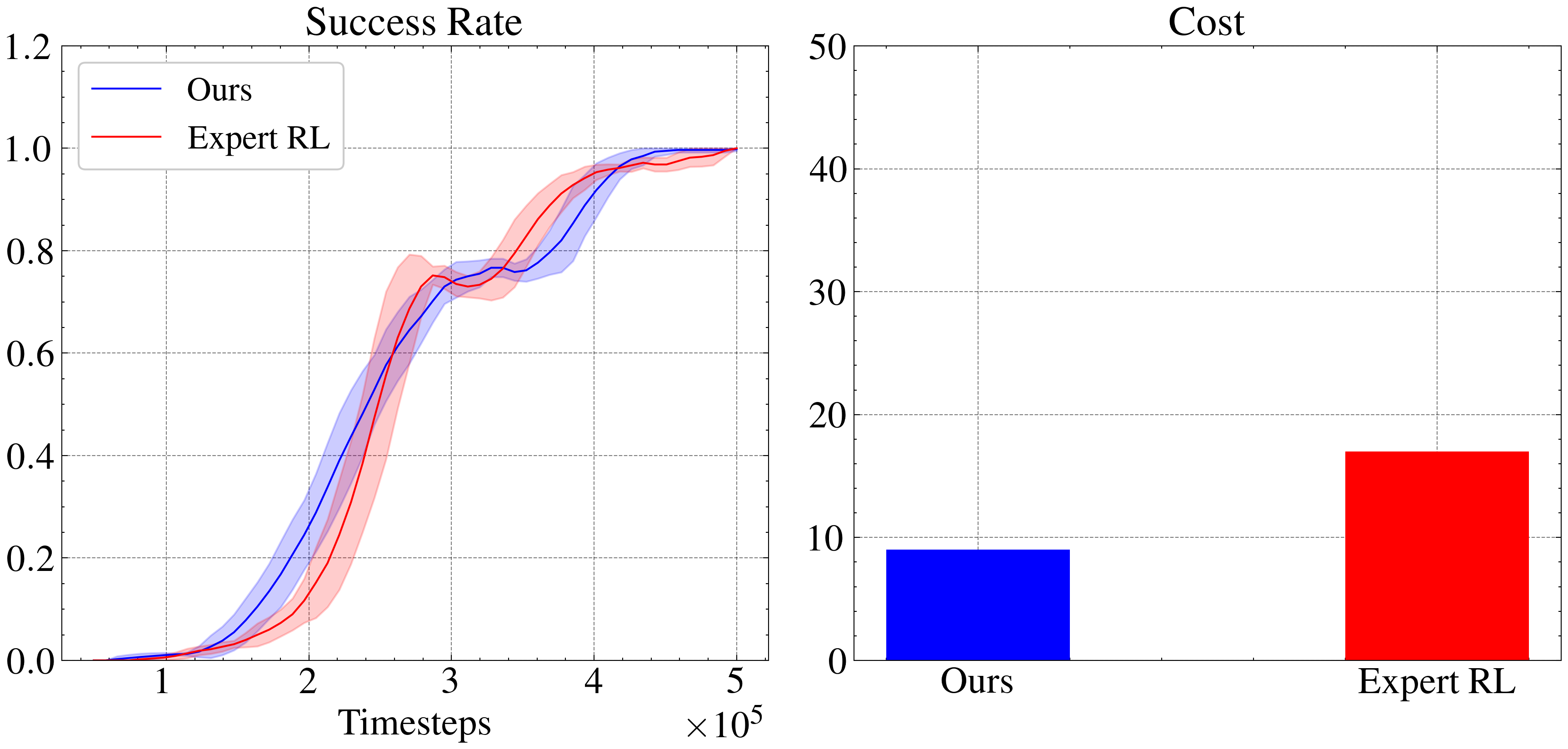}}
    \caption{Comparisons between the proposed framework and RL baselines in the     Safety-Gymnasium reach-avoid tasks.}
    \label{fig:safety-gym-comparisons}
\end{figure}

A single agent following Ackermann\footref{fn:kinematic-models} kinematics must reach a goal location while avoiding hard and soft obstacle regions. The task is described by the TBT specification
\[
\tree \defeq \Parallel_2(\ev\alw \mathit{reachGoal}, \alw \mathit{avoidCollision})
\]
where the predicate $\mathit{reachGoal}$ measures the Euclidean distance between the agent's current location and the goal position, and $\mathit{avoidCollision}$ indicates whether the distance between the agent and an obstacle is less than a defined threshold. The observation space is $60$-dimensional, comprising lidar distance measurements to obstacles and goals as well as proprioceptive features such as the agent's pose and velocity, with noise implicitly applied to the lidar measurements. Although the TBT repair tool uses a Dubins vehicle model for the LP formulation, the extracted potential functions and RL training generalize to the Ackermann agent, as both models share a similar 2D pose space and kinematic structure. Note that due to randomized environment configurations, traversal of some obstacle regions may be unavoidable in certain episodes. Both methods achieved $100\%$ task success rates with similar cost rates (\autoref{fig:single-comparison}), though the proposed approach converged marginally faster, suggesting that the repair-based potential provides a useful early guidance signal even under configuration randomization.

\subsubsection{Multi-Agent Reach-Avoid}

Two robots following unicycle\footref{fn:kinematic-models} kinematics must collectively cover two goal locations, with exactly one robot per goal at termination, while avoiding obstacles. The task is described by the TBT specification
\[
\tree \defeq \Parallel_3(\ev\alw \mathit{reachGoal_1}, \ev\alw \mathit{reachGoal_2}, \alw \mathit{avoidCollision})
\]
The predicates $\mathit{reachGoal_1}$ and $\mathit{reachGoal_2}$, represent the minimum distance between each agent and the two goals, while $\mathit{avoidCollision}$ is applied similar to the single-agent case for each agent. The observation space has a total of $152$ dimensions ($76$ per robot), with noise implicitly applied to the lidar measurements. As in the single-agent case, the Dubins-based repair generalizes to the unicycle agent due to their shared 2D pose representation, without requiring any modification to the reward shaping pipeline. Both methods achieved $100\%$ task success rates; however, the proposed approach incurred a $47\%$ lower cost rate than the expert-reward baseline (\autoref{fig:multi-comparison}). This is a significant result: despite the potential function being inferred from a limited, configuration-agnostic set of repaired demonstrations, it provides a meaningful guidance signal that generalizes to unseen configurations and multi-agent interactions, reducing obstacle violations without sacrificing task performance.

\subsubsection{Discussion}

\begin{table}[htbp]
\centering
\caption{PPO Hyperparameters for grid-world, safety-gymnasium single-agent (SG-Single), and multi-agent (SG-Multi). Unlisted hyperparameters are set to the default values in~\cite{stable_baselines3}; the neural networks are all fully connected.}
\label{tab:hyperparams}
\begin{tabular}{lccc}
\toprule
\textbf{Hyperparameter} & \textbf{Grid-World} & \textbf{SG-Single} & \textbf{SG-Multi}\\
\midrule
Horizon & 300 & 1000 & 1000 \\
Learning rate & $10^{-3}$ & $3 \times 10^{-4}$ & $3 \times 10^{-4}$\\
Discount factor $\gamma$ & $0.95$ & $0.99$ & $0.99$ \\
Target KL & $-$ & $0.2$ & $0.2$ \\
Batch size & $64$ & $256$ & $256$ \\
\# Epochs & $4$ & $10$ & $10$ \\
Steps per update & $1024$ & $8192$ & $8192$ \\
Network architecture & $[64, 64]$ & $[256, 256]$ & $[256, 256]$ \\
Activation function & $\mathtt{Tanh}$ & $\mathtt{ReLU}$ & $\mathtt{ReLU}$ \\
Training timesteps & $10^{4}$ & $2.5 \times 10^{5}$ & $5 \times 10^{5}$\\
$\alpha, \beta$ & $0.5, 0.5$ & $0.4, 0.2$ & $0.4, 0.2$ \\
\bottomrule
\end{tabular}
\end{table}

The environment and learning configurations used in our experiments are listed in~\autoref{tab:hyperparams}. Collectively, these experiments highlight several important properties of the proposed framework. Despite the significant increase in observation dimensionality and stochasticity relative to the grid-world, the framework requires no architectural modifications, as the potential functions operate purely on the agent's 2D pose and are decoupled from the full observation space. Moreover, the use of a Dubins model in the TBT repair tool generalizes across kinematically similar mobile robot platforms (Ackermann in the single-agent case and unicycle in the multi-agent case) demonstrating that the repair-based potential functions are not tied to a specific kinematic model. \emph{By decoupling the potential function from both the observation space and the agent's kinematic model, the proposed framework scales naturally to high-dimensional, noisy, and multi-agent settings without additional engineering effort.}

Despite promising results, the proposed framework has several limitations. The TBT repair tool relies on an approximate kinematic model and operates offline, which restricts its use in real-time or model-free settings. Its configuration-agnostic potential functions can also be imprecise in environments with randomized obstacles and goals, where a repaired pose from one configuration may conflict with another. Future work includes conditioning potentials on current observations, extending repair to higher-dimensional systems, and handling partially known or data-inferred TBT specifications.


\section{Conclusion}
\label{sec:conclusion}

	In this paper, we presented a formal framework for robot policy learning that integrates TBT-based trajectory repair with reinforcement learning, enabling policy acquisition from imperfect demonstrations using only an approximate kinematic model for repair. Potential functions extracted from the repaired dataset shape the reward signal for RL, providing a dense and specification-consistent guidance signal that is decoupled from the full observation space and generalizes across similar robotic platforms. Experiments on discrete grid-world navigation and continuous single and multi-agent reach-avoid tasks demonstrate that the proposed approach consistently matches or exceeds expert-designed reward baselines while incurring significantly lower obstacle cost rates.


\clearpage
\acknowledgments{This work was partially supported by the Army Research Laboratory Cooperative Agreement No. W911NF-23-2-0040, the NSF EFRI 2422282, a Northrop Grumman Corporation grant, the Lockheed Martin Chair in Systems Engineering, and the Brendan Iribe Endowed Professorship at the University of Maryland.}


\bibliography{refs}

@misc{schirmer2025_repairtbt,
      title={Trace Repair for Temporal Behavior Trees}, 
      author={Sebastian Schirmer and Philipp Schitz and Johann C. Dauer and Bernd Finkbeiner and Sriram Sankaranarayanan},
      year={2025},
      eprint={2509.08610},
      archivePrefix={arXiv},
      primaryClass={cs.LO},
}

@inproceedings{schirmer2024_tbt,
author = {Schirmer, Sebastian and Singh, Jasdeep and Jensen, Emily and Dauer, Johann and Finkbeiner, Bernd and Sankaranarayanan, Sriram},
title = {Temporal Behavior Trees: Robustness and Segmentation},
year = {2024},
isbn = {9798400705229},
publisher = {Association for Computing Machinery},
address = {New York, NY, USA},
doi = {10.1145/3641513.3650180},
booktitle = {Proceedings of the 27th ACM International Conference on Hybrid Systems: Computation and Control},
articleno = {9},
numpages = {14},
location = {Hong Kong SAR, China},
series = {HSCC '24}
}

@INPROCEEDINGS{matheu2025_omtbt,
  author={Matheu, Ryan and Puranic, Aniruddh G. and Baras, John S. and Belta, Calin},
  booktitle={2025 European Control Conference (ECC)}, 
  title={OMTBT: Online Monitoring of Temporal Behavior Trees with Applications to Closed-Loop Learning}, 
  year={2025},
  volume={},
  number={},
  pages={2129-2135},
  doi={10.23919/ECC65951.2025.11187275}
  }

@inproceedings{matheu2025_bt2automata,
author = {Matheu, Ryan and Puranic, Aniruddh G. and Baras, John S. and Belta, Calin},
title = {BT2Automata: Expressing Behavior Trees as Automata for Formal Control Synthesis},
year = {2025},
isbn = {9798400715044},
publisher = {Association for Computing Machinery},
address = {New York, NY, USA},
doi = {10.1145/3716863.3718042},
booktitle = {Proceedings of the 28th ACM International Conference on Hybrid Systems: Computation and Control},
articleno = {13},
numpages = {11},
location = {Irvine, CA, USA},
series = {HSCC '25}
}

@article{ppo_schulman,
  author    = {John Schulman and
               Filip Wolski and
               Prafulla Dhariwal and
               Alec Radford and
               Oleg Klimov},
  title     = {Proximal Policy Optimization Algorithms},
  journal   = {CoRR},
  volume    = {abs/1707.06347},
  year      = {2017},
  eprinttype = {arXiv},
  eprint    = {1707.06347},
}

@article{stable_baselines3,
  author  = {Antonin Raffin and Ashley Hill and Adam Gleave and Anssi Kanervisto and Maximilian Ernestus and Noah Dormann},
  title   = {Stable-Baselines3: Reliable Reinforcement Learning Implementations},
  journal = {Journal of Machine Learning Research},
  year    = {2021},
  volume  = {22},
  number  = {268},
  pages   = {1--8},
}

@misc{gym_simplegrid,
  author = {Leo D'Amato},
  title = {SimpleGrid: Simple Grid Environment for Gymnasium},
  year = {2022},
  publisher = {GitHub},
  journal = {GitHub repository},
  howpublished = {\url{https://github.com/damat-le/gym-simplegrid}},
}

@inproceedings{donze_robust_2010,
  author    = {Alexandre Donz{\'{e}} and
               Oded Maler},
  title     = {Robust Satisfaction of Temporal Logic over Real-Valued Signals},
  booktitle = {FORMATS},
  year      = {2010},
}

@article{fainekos_robustness_2009,
  author    = {Georgios E. Fainekos and
               George J. Pappas},
  title     = {Robustness of temporal logic specifications for continuous-time signals},
  journal   = {Theoretical Computer Science},
  year      = {2009},
}

@inproceedings{ng_policy_invariance,
  author    = {Andrew Y. Ng and
               Daishi Harada and
               Stuart Russell},
  title     = {Policy Invariance Under Reward Transformations: Theory and Application
               to Reward Shaping},
  booktitle = {Proceedings of the Sixteenth International Conference on Machine Learning
               {(ICML} 1999), Bled, Slovenia, June 27 - 30, 1999},
  pages     = {278--287},
  publisher = {Morgan Kaufmann},
  year      = {1999},
}

@article{bentley1975multidimensional,
author = {Bentley, Jon Louis},
title = {Multidimensional binary search trees used for associative searching},
year = {1975},
issue_date = {Sept. 1975},
publisher = {Association for Computing Machinery},
address = {New York, NY, USA},
volume = {18},
number = {9},
issn = {0001-0782},
doi = {10.1145/361002.361007},
journal = {Commun. ACM},
month = sep,
pages = {509--517},
numpages = {9}
}

@inproceedings{ji2023safetygym,
  title={Safety Gymnasium: A Unified Safe Reinforcement Learning Benchmark},
  author={Jiaming Ji and Borong Zhang and Jiayi Zhou and Xuehai Pan and Weidong Huang and Ruiyang Sun and Yiran Geng and Yifan Zhong and Josef Dai and Yaodong Yang},
  booktitle={Thirty-seventh Conference on Neural Information Processing Systems Datasets and Benchmarks Track},
  year={2023},
}

@inproceedings{PDN_ALSTL,
  author = {Puranic, Aniruddh G. and Deshmukh, Jyotirmoy V. and Nikolaidis, Stefanos},
  booktitle = {2024 IEEE/RSJ International Conference on Intelligent Robots and Systems (IROS)},
  title = {Signal Temporal Logic-Guided Apprenticeship Learning},
  year = {2024},
  volume = {},
  number = {},
  pages = {11147-11154},
  doi = {10.1109/IROS58592.2024.10801924},
}

@article{PDN21,
  author = {Puranic, Aniruddh G. and Deshmukh, Jyotirmoy V. and Nikolaidis, Stefanos},
  journal = {IEEE Robotics and Automation Letters (RA-L). Presented at IROS},
  title = {Learning From Demonstrations Using Signal Temporal Logic in Stochastic and Continuous Domains},
  year = {2021},
  volume = {6},
  number = {4},
  pages = {6250-6257},
  doi = {10.1109/LRA.2021.3092676}
}

@inproceedings{aksaray_qlearning_2016,
  author    = {Derya Aksaray and
               Austin Jones and
               Zhaodan Kong and
               Mac Schwager and
               Calin Belta},
  title     = {Q-Learning for robust satisfaction of signal temporal logic specifications},
  booktitle = {55th {IEEE} Conference on Decision and Control, {CDC} 2016, Las Vegas,
               NV, USA, December 12-14, 2016},
  pages     = {6565--6570},
  publisher = {{IEEE}},
  year      = {2016},
  doi       = {10.1109/CDC.2016.7799279},
}

@inproceedings{li_reinforcement_2017,
  author    = {Xiao Li and
               Cristian Ioan Vasile and
               Calin Belta},
  title     = {Reinforcement learning with temporal logic rewards},
  booktitle = {2017 {IEEE/RSJ} International Conference on Intelligent Robots and
               Systems, {IROS} 2017, Vancouver, BC, Canada, September 24-28, 2017},
  pages     = {3834--3839},
  publisher = {{IEEE}},
  year      = {2017},
  doi       = {10.1109/IROS.2017.8206234},
}

@inproceedings{anandb_rl,
  author    = {Anand Balakrishnan and
               Jyotirmoy V. Deshmukh},
  title     = {Structured Reward Shaping using Signal Temporal Logic specifications},
  booktitle = {2019 {IEEE/RSJ} International Conference on Intelligent Robots and
               Systems, {IROS} 2019, Macau, SAR, China, November 3-8, 2019},
  pages     = {3481--3486},
  publisher = {{IEEE}},
  year      = {2019},
  doi       = {10.1109/IROS40897.2019.8968254},
}

@article{jiang2020temporallogicbased,
    title={Temporal-Logic-Based Reward Shaping for Continuing Reinforcement Learning Tasks},
    volume={35},
    DOI={10.1609/aaai.v35i9.16975},
    number={9},
    journal={Proceedings of the AAAI Conference on Artificial Intelligence},
    author={Jiang, Yuqian and Bharadwaj, Suda and Wu, Bo and Shah, Rishi and Topcu, Ufuk and Stone, Peter}, year={2021},
    month={May},
    pages={7995-8003}
}

@inproceedings{wenchao_safety_al,
  author    = {Weichao Zhou and
               Wenchao Li},
  title     = {Safety-Aware Apprenticeship Learning},
  booktitle = {Computer Aided Verification - 30th International Conference, {CAV}
               2018, Held as Part of the Federated Logic Conference, FloC 2018, Oxford,
               UK, July 14-17, 2018, Proceedings, Part {I}},
  series    = {Lecture Notes in Computer Science},
  volume    = {10981},
  pages     = {662--680},
  publisher = {Springer},
  year      = {2018},
  doi       = {10.1007/978-3-319-96145-3\_38},
}

@inproceedings{InnesRSS20,
  author    = {Craig Innes and
               Subramanian Ramamoorthy},
  title     = {Elaborating on Learned Demonstrations with Temporal Logic Specifications},
  booktitle = {Robotics: Science and Systems XVI, Virtual Event / Corvalis, Oregon,
               USA, July 12-16, 2020},
  year      = {2020},
  doi       = {10.15607/RSS.2020.XVI.004},
}

@inproceedings{cho_mpc_stl,
  author    = {Kyunghoon Cho and
               Songhwai Oh},
  title     = {Learning-Based Model Predictive Control Under Signal Temporal Logic
               Specifications},
  booktitle = {2018 {IEEE} International Conference on Robotics and Automation, {ICRA}
               2018, Brisbane, Australia, May 21-25, 2018},
  pages     = {7322--7329},
  publisher = {{IEEE}},
  year      = {2018},
  doi       = {10.1109/ICRA.2018.8460811},
}

@inproceedings{wen_lfd_highlevel,
  author    = {Min Wen and
               Ivan Papusha and
               Ufuk Topcu},
  title     = {Learning from Demonstrations with High-Level Side Information},
  booktitle = {Proceedings of the Twenty-Sixth International Joint Conference on
               Artificial Intelligence, {IJCAI} 2017, Melbourne, Australia, August
               19-25, 2017},
  pages     = {3055--3061},
  publisher = {ijcai.org},
  year      = {2017},
  doi       = {10.24963/ijcai.2017/426},
}

@article{belta_lfd,
  author    = {Xiao Li and
               Yao Ma and
               Calin Belta},
  title     = {Automata Guided Reinforcement Learning With Demonstrations},
  journal   = {CoRR},
  volume    = {abs/1809.06305},
  year      = {2018},
  eprinttype = {arXiv},
  eprint    = {1809.06305},
}

@article{ravichandar_recent_2020,
  author  = {Ravichandar, Harish and Polydoros, Athanasios S. and Chernova, Sonia and Billard, Aude},
  title   = {Recent Advances in Robot Learning from Demonstration},
  journal = {Annual Review of Control, Robotics, and Autonomous Systems},
  volume  = {3},
  number  = {1},
  pages   = {297-330},
  year    = {2020},
}

@misc{gurobi,
  author = {{Gurobi Optimization, LLC}},
  title = {{Gurobi Optimizer Reference Manual}},
  year = 2026,
  url = "https://www.gurobi.com"
}

\end{document}